%% file: plain.tex
\documentclass[eng]{IEEEtran}
\pdfoutput=1
\usepackage[english]{babel}
\usepackage{makecell}
\usepackage{algorithmic}
\usepackage{amsmath,amssymb,amsfonts}
\usepackage{graphicx}
\usepackage{booktabs}
\usepackage[labelfont=bf]{caption}
\usepackage{caption}
\usepackage{subcaption}
\usepackage{hyperref}
\usepackage{xcolor}
\newcommand{\e}[1]{\exp\left(#1\right)}

\usepackage[style=ieee, sorting=none]{biblatex}
\title{\title{Flexible framework for generating synthetic electrocardiograms and photoplethysmograms }}
\author{\IEEEauthorblockN{Katri Karhinoja\textsuperscript{1}, Antti Vasankari\textsuperscript{1}, Jukka-Pekka Sirkiä\textsuperscript{1}, Antti Airola\textsuperscript{1}, David Wong\textsuperscript{2} \\ and Matti Kaisti\textsuperscript{1}\IEEEauthorrefmark{1}}

\IEEEauthorblockA{\textsuperscript{1} Department of Computing, Faculty of Technology, University of Turku, Turku 20520, Finland}\\
\IEEEauthorblockA{\textsuperscript{2} Leeds Institute of Health Sciences, School of Medicine, University of Leeds, Leeds LS2 9LH, United Kingdom
}
\thanks{\IEEEauthorrefmark{1}Corresponding author:mkaist@utu.fi }}

\begin{document}

\renewcommand{\abstract}[1]
{
  \small
  \textbf{\textit{Abstract---}} #1
}
\maketitle
\abstract{\textbf{By generating synthetic biosignals, the quantity and variety of health data can be increased. This is especially useful when training machine learning models by enabling data augmentation and introduction of more physiologically plausible variation to the data. For these purposes, we have developed a synthetic biosignal model for two signal modalities, electrocardiography (ECG) and photoplethysmography (PPG). The model produces realistic signals that account for physiological effects such as breathing modulation and changes in heart rate due to physical stress. Arrhythmic signals can be generated with beat intervals extracted from real measurements. The model also includes a flexible approach to adding different kinds of noise and signal artifacts. The noise is generated from power spectral densities extracted from both measured noisy signals and modeled power spectra. Importantly, the model also automatically produces labels for noise, segmentation (e.g. P and T waves, QRS complex, for electrocardiograms), and artifacts. We assessed how this comprehensive model can be used in practice to improve the performance of models trained on ECG or PPG data. For example, we trained an LSTM to detect ECG R-peaks using both real ECG signals from the MIT-BIH arrythmia set and our new generator. The F1 score of the model was 0.83 using real data, in comparison to 0.98 using our generator. In addition, the model can be used for example in signal segmentation, quality detection and bench-marking detection algorithms. The model code has been released in \url{https://github.com/UTU-Health-Research/framework_for_synthetic_biosignals}.}}

\providecommand{\keywords}[1]
{
  \small
  \textbf{\textit{Keywords---}} #1
}

\section{Introduction}
\label{sec:introduction}

Analysis of physiological time series plays a crucial role in understanding physiological processes, as well as in diagnosing and monitoring various health conditions. The advent of wearable devices and remote monitoring systems has greatly expanded the availability of physiological time series data. However, this so-called "free-living" data is typically unlabeled, limiting its use in data driven applications \cite{diaz2022data} such as cardiac cycle segmentation. In addition, increasing privacy concerns greatly hinder the collection, availability and sharing of health data \cite{mcgraw2021privacy,cohen2019big,price2019privacy}. One approach to address these issues could be synthetic data defined by James Jordon as "data that has been generated using a purpose built mathematical model or algorithm, with the aim of solving a (set of) data science task(s)" \cite{jordon2022synthetic}. 

There are, however, certain limitations to synthetic signals. It can be difficult to accurately mimic real-world signals due to the complex physiological processes. Validation of synthetic signals can be challenging, as there is often a lack of ground truth data for comparison. While synthetic biosignals can be designed to cover a wide range of scenarios, they may not accurately reflect all real-world conditions.

Despite the limitations to the accuracy and generalizability, generating synthetic biosignals is an important step in developing and evaluating deep learning models in wearable and telemonitoring systems as they can provide a controlled and flexible source of data for training and testing algorithms. By using synthetic biosignals for anomaly detection, the accuracy, reliability and interpretability of the model could be improved. Additionally, synthetic health data could be used for e.g. algorithm benchmarking, studying effect of sampling rate, and educational purposes. 

In this study we extend previously presented synthetic dynamic electrocardiogram (ECG) model \cite{kaisti2023domain} and photoplethysmogram (PPG) model \cite{sirkia2024wearable}. The presented model includes i) arbitrary noise generation including stationary and non-stationary noises, ii) comprehensive beat-interval generation including long-term beat-interval correlations and iii) step like change in the mean heart rate in the interval array and iv) the possibility of using annotated atrial fibrillation intervals, v) ability to generate both ECG and PPG waveforms, vi) the generation of longitudinal signal with user specified properties and vii) a domain randomization scheme allowing the generation of unique biosignal realizations for testing algorithms and training models. The signals include labeling information for each beat as well as a quantitative quality information. The signals can be generated to have significant variability due to domain randomization, and thus the signals can resemble various real conditions of physiological origin. The domain randomization has also been shown to improve model robustness \cite{kaisti2023domain}.

The major advantage of the proposed framework is that it is the most comprehensive publicly available synthetic signal generator. It includes flexibility over noise with both measured and modeled noise as well as addition of artifacts. The model is multimodal with ability to generate both ECG and PPG signals. The model could also be further extended to cover other modalities such as ballistocardiograms. In addition there is domain randomization over noise, signal parameters and beat intervals. 

\begin{figure*}[]
  \centering  \includegraphics[width=0.9\textwidth]{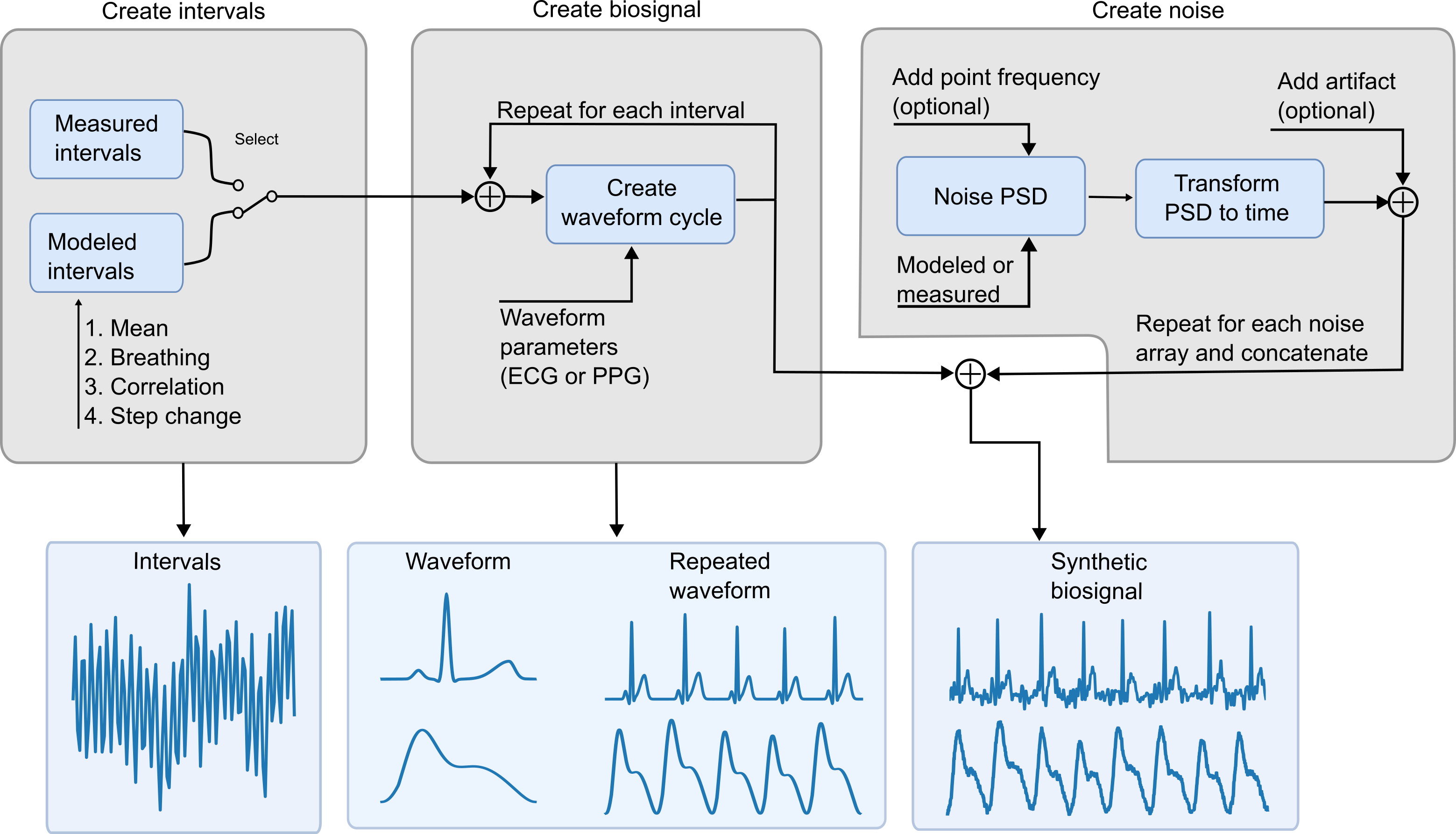}
  \caption{Block diagram of the biosignal generation model. The model consists of three parts: beat interval generator, signal generator and noise generator. The beat intervals are affected by mean heart rate, respiratory modulation, long-term correlation and a change in the mean heart rate. The biosignals (ECG or PPG) are generated from the beat intervals with given parameters (amplitude, location and width) to each wave (e.g. ECG T wave). The noise generator produces time domain noise from noise PSDs and optionally adds artifacts to the noise. Finally, the signal and noise are combined into a realistic synthetic biosignal.}  
  \label{fig:block}
\end{figure*}

\section{Related work}

The existing approaches to generate synthetic biosignals can be categorized to parametric, physiological and deep learning models. 

With parametric models, the output signal is generated by a mathematical model that is controlled by the input parameters. One of the most used parametric models for ECG is the dynamic model that is based on three coupled differential equations \cite{mcsharry2003dynamical}. Other parametric ECG models have used trigonometric and linear functions, and Fourier series to generated ECG signals \cite{dolinsky2018model} \cite{kubicek2014design}. Parametric models for PPG signals also exist, including stochastic model that synthesizes signals based on one analyzed signal \cite{martin2013stochastic}, and model for irregular PPG generation that combines two Gaussian functions \cite{tang2020ppgsynth}. 

Physiological models work by computationally modeling a physiological system to generate realistic biosignals. ECG signals can be simulated with computational finite element whole heart models \cite{sovilj2014simplified,neic2017efficient} that model the electrical activity of the heart. In addition, cardiovascular system has been computationally modeled to synthesize healthy and pathological PPG signals \cite{mazumder2022synthetic}. Compared to parametric models, physiological models are able to better consider physiological variations. However, physiological models are computationally heavy and complex and randomization of the signals is limited.

Deep learning models, such as generative adversarial networks \cite{qin2023novel,adib2021synthetic,hazra2020synsiggan,piacentino2021generating,adib2023synthetic,wulan2020generating}, can be trained on data measured from real patients to generate synthetic biosignal data with similar characteristics. A limitation of the approach is the need to have large and diverse enough training data that accurately covers the full spectrum of variability found in biosignals. In addition, it has been shown that generative models can be vulnerable to attacks by malicious users that aim to infer properties of sensitive training data, such as the presence of a certain individual in the training set \cite{chen2020gan}.

Compared to physiological and deep learning models, parametric models are better to create controlled variation to the signals. However, the ability of the model to generate variability to the signals is limited by the parameters. In addition, biosignals are complex signals originating from several physiological processes so taking into account all the affecting factors is difficult with a parametric model.

\section{Synthetic model}
The synthetic biosignal model consists of three parts: beat interval generation model, signal generation model and noise generation model. The beat intervals, generated for the signal generation model, are affected by average heart rate, breathing modulation, long-term correlation and if selected by a step change in the average heart rate. Optionally, the user can bypass the beat interval generation process by giving an array of existing beat intervals. As a special case, the user can choose to use the beat intervals from \cite{moody2001impact} which is implemented in the code. The synthetic signal generation process is controlled by parameters defining the locations, amplitudes and widths for each wave (e.g. the T wave of an ECG signal). All these parameters can be randomized. For ECG, the randomization is done independently for each parameter whereas for PPG the randomization is done jointly for all the parameters to ensure that the PPG signal remains physiological. The noise is generated from a power spectral density (PSD) which can be either from a real measurement or mathematically modeled. In addition, a point frequency can be added to the PSD to model the mains interference. After generating the time domain noise from the PSD, an artifact extracted from the signals in the MIT-BIH Noise Stress Test Database \cite{moody1984bih} can be added. The noise is also generated in a randomized manner. Finally, the synthetic biosignal and the noise array are added together, resulting in a noisy biosignal. The block diagram of the model is shown in Fig \ref{fig:block}.

\subsection{Beat-interval generation}

\subsubsection{Beat-interval model}
Heartbeat intervals were modeled as
\begin{equation}\label{beatinterval}
    \theta_i:=\mu+\beta\sin(2\pi f_b t_{i-1})+\gamma_i,
\end{equation}
in which $\mu$ is the average beat interval, $\beta$ is a breathing coefficient $t_{i-1}$ is the sum of previous intervals 
\begin{equation*}
    t_{i-1}=\sum_{j=1}^{i-1}\theta_j, \; \forall i\geq2 \; \text{and} \; t_0 = 0,
\end{equation*}
representing time before $i$-th beat, therefore the function $\beta \sin(2\pi f_bt_{i-1})$ oscillates between $-\beta$ and $\beta$, and breathing frequency $f_b$ determines the number of cycles occurring in one unit of time. Parameter $\gamma_i$ is a stochastic component from the presentation in article \cite{kantelhardt2003transiencorr}, that is slightly modified, as shown in appendix \ref{modifiedgamma}, that includes fluctuation with transient correlations akin to an observation of a healthy, awake subject. The correspondence is validated using detrended fluctuation analysis (DFA). Breathing component generates the respiratory modulation of the heart rate to the beat intervals.

\subsubsection{Step change in average heart rate}

There can also be a change in the average heart rate which is generated by multiplying the beat intervals with a sigmoid function
\begin{equation}\label{sigmoid}
    S:\mathbb{R}\to(0,1),\text{ where }S(x)=\frac{1}{1+\e{-x}},
\end{equation}
so the beat intervals gradually increase or decrease. From the beat interval model \eqref{beatinterval} the change in heart rate is created by the mean of the beat intervals $\mu$ to a new mean $\mu' \neq \mu$ in the following manner, that is more precisely justified in the appendix \ref{sigmoidoversignal}:
\begin{equation*}
    \theta_i:=\mu+(\mu'-\mu)S\left (\frac{x-d}{\tau}\right )+\epsilon_i, 
\end{equation*}
in which $\epsilon_i=\beta\sin(2\pi f_b t_i)+\gamma_i$ is the variation in the model \eqref{beatinterval}, parameter $d=\frac{lT}{\mu}$, where $l\in[0,1]$ and $T$ is the length of the signal, determines the middle point of the transition and parameter $\frac{1}{\tau} \in [0.1,1]$ is the gradient of the step.
The coefficient $\tau$ is defined, such that over five beat intervals, with an input value of $1$, $99.5\%$ of the change occurs, with $2$, $90.5\%$ of the change occurs, with $3$, $76.2\%$ of the change occurs, and so forth. 

\subsubsection{Beat-intervals from measurements}
Beat intervals from real measurements can also be given to the model to allow generation of synthetic biosignals with intervals extracted from e.g. atrial fibrillation recordings. The developed computer code has a built-in functionality to select beat intervals extracted from the MIT-BIH Arrhythmia Database \cite{moody2001impact,goldberger2000physiobank}. 

\subsection{Signal generation}

\subsubsection{Waveform model}

Let's assume that each ECG cycle consists of five waves (P, Q, R, S and T) and each PPG cycle consists of two waves (systole and diastole) that are all similar to the Gaussian function
\begin{equation}\label{gaussian}
    a\e{-\frac{(x-\mu)^2}{2w^2}}.
\end{equation}
The parameter $w$ affects the width of the waves, and the highest point of the curve is clearly where $x=\mu$, thus parameter $a$ represents the amplitude.

Because each Gaussian wave is calculated separately, in order to avoid discontinuity in the derivatives and for the beat intervals to affect the amplitude of the synthetic biosignal, the waves are created and summed utilizing the derivatives of the function \eqref{gaussian}.

Consider a phase function
\begin{equation*}
    \phi: [0,1] \to [-\pi, \pi], \; \phi(x)=2\pi x-\pi,\text{ where }x: \mathbb{R}_{\geq 0} \to [0, 1]
\end{equation*}
is an irregular sawtooth wave of time, as defined in appendix \ref{xoftime}, where the width of each ramp is determined by the sequence $(\theta_k)_{k=1}^n$, that represents the beat intervals. Therefore, function $\phi$ depicts each cardiac cycle as a single rotation around a circle.

Furthermore, to generate asymmetry for the wave, parameter $m$ is added. With these modifications to Gaussian function \eqref{gaussian}, the partial derivative can be conducted as
\begin{equation}\label{dermodel}
    g(\phi) = \frac{\partial}{\partial x} a\e{-\frac{m\phi^2}{2w^2}} = -\frac{2\pi m a \phi}{w^2}\e{-\frac{m\phi^2}{2w^2}},
\end{equation}
where $m \neq 1$ for non-zero skewness. For ECG, in the case of T-wave that is generally left skewed, the asymmetric characteristic is generated via values $m > 1$, when $\phi > 0$.

A phase function $\phi_j$ is created for each wave $j$. The time difference between the waves is done by shifting the zero of the function to the location of each peak as $\phi_j=(2x-1+d_j)\pi$. The derivative of the signal is formed utilizing these phase functions in the equation \eqref{dermodel}. The functions $g(\phi_j)$ are summed and integrated over time leading to the final signal as detailed more extensively in the appendix \ref{sumofwaves}.

Concisely, derivative of the biosignal for a single cardiac cycle, as determined by the wave parameters, is created around a circle. The circumference is determined by the length of the beat interval, or in discrete time, beat interval establishes the number of samples around the circle. To achieve the complete signal, the derivative is numerically integrated using trapezoid method over every sample interval. Therefore, the longer the beat interval, the longer the distances between the waves, and the higher their amplitudes and widths. Furthermore, using the integral smooths over the possible baseline differences and is better at representing a continuously differentiable function.

\subsection{Noise generation}
\label{sec:algorithms}

\subsubsection{Stationary noise}

To generate noise from real signals, PPG measurements were taken with the subject walking or moving the hand with the PPG device \cite{sirkia2024wearable}. In addition, noise measurements containing baseline wander and muscle artifacts were downloaded from MIT-BIH Noise Stress Test Database \cite{moody1984bih, goldberger2000physiobank}. PSDs were computed from the measurements using the Welch's method \cite{welch1967use}.

\subsubsection{Noise model}

The mathematical PSD noise model was a combination of $1/f$ noise and white noise defined as
\begin{equation}\label{psd}
    PSD(f)=\frac{1}{f^\alpha}+\sigma^2,
\end{equation}
where constant $\alpha \in [0, 5]$ emphasizes low frequencies and the uniformly distributed white noise is denoted as $\sigma^2 \in [0, 2]$.
The first part of the model \eqref{psd} is then normalized by dividing by its average $\bar{P}$.
Then a coefficient $c \in [0, 2]$ is assigned to determine the overall effect of the normalized $1/f$ noise such that the model PSD is redefined as:
\begin{equation}\label{modelPSD}
    PSD(f):=\frac{c}{f^\alpha}/\bar{P}+\sigma^2.
\end{equation}
Therefore, the total power of $1/f$ noise and white noise can be adjusted independently.

Consider a sequence $(f_k)_{k=1}^n=(f_1,...,f_n), \; 0<f_j<f_{j+1} \; \forall j,$ to represent the positive frequency intervals of the PSD (model or from real measurement) and the corresponding powers are denoted as $(PSD_k)_{k=1}^n=(PSD(f_1),...,PSD(f_n))$. To randomize the values $(PSD_k)_{k=1}^n$, each of them is multiplied by a complex Gaussian random variable from sequence $(z_k)_{k=1}^n$, where $z_j = a_j + b_j i \in \mathbb{C}$ and $a_j, b_j \sim N(0, 1) \; \forall j$. Let's denote the randomized square root of $PSD$ as
\begin{equation*}
    R_i:=\frac{\sqrt{PSD_iz_i}}{2}.
\end{equation*}
In order to achieve a real valued sequence with inverse fast Fourier transform (IFFT) of the PSD, the randomized PSD, $(R_k)_{k=1}^n$, has to be symmetric with respect to the origin for its imaginary part. Therefore,
\begin{align*}
    R(-f_j)&:=R(f_j)^* \; \forall f_j, \text{ where } R^* \text{ is the complex conjugate}, \text{ and}\\
    R(0)&:=0 \text{ is the direct current component fixed to zero} \\
    &\Rightarrow IFFT\left((R_k)_{k=-n}^n\right)=(R'_k)_{k=1}^n, \; R'_i\in\mathbb{R} \; \forall i.
\end{align*}
Thus, noise as a real-valued signal $(R_k')_{k=1}^n$ in time domain is achieved from the model PSD \eqref{modelPSD}. In practice, due to rounding error, there is a residual imaginary part that can be ignored.

\subsubsection{Artifact augmentation}
The noisy synthetic signals can be augmented by adding an artifact. Baseline wander and muscle artifact noises are loaded from the MIT-BIH Noise Stress Test database \cite{moody1984bih, goldberger2000physiobank}. Random segment of either of the noise signals are obtained, normalized to the range of [0, 1] and multiplied by a uniformly distributed random number between [0.1, 1] to vary the strength of the artifact. The resulting artifact is then added to the [0, 1] normalized final signal.

\subsection{Randomization}

To produce a set of random signals, the generation process of beat intervals, synthetic signals and noises were randomized. For the beat intervals the average heart rate was randomized. In addition, a step change in the heart rate was randomly added to the beat intervals.

For the signal, the amplitudes, widths and locations of the waves (e.g. P, Q, R, S and T for ECG) were randomized. Regarding ECG also the symmetry of the T wave was randomized. For ECG the randomization was done independently whereas for PPG, amplitude, width and location of the waves were randomized by multiplying the randomization range with the same uniformly distributed number between [0,1] and adding that to the lower limit. This ensures that the generated PPG remains physiologically realistic. 

Time domain noise was randomized by generating the time domain array from a randomized power spectrum. In addition, the parameters of the noise were randomized including the type of the noise (walking, hand movement, muscle artifact, baseline wander or model PSD) and amplitude of the noise. The time domain noise array was standardized by dividing them with standard deviation and then multiplied with the randomized amplitude. Also, point frequency, randomly chosen between 0 and Nyquist frequency, was added to the PSD before converting it to time domain. An artifact was also randomly added to the noise. For the model PSD, the exponent $\alpha$, constant $c$ and power of the white noise were randomized. The model PSD was also normalized by dividing it with its average power. Randomization limits are shown in table \ref{tab:modpar}. Randomization of all the parameters were done by drawing uniformly distributed samples over the randomization range.

\subsection{Long-term signals}

Longitudinal signals where the noise properties change during the signals in terms of noise type or power can be generated by concatenating multiple noise arrays. The time-domain noise arrays were concatenated by reducing the end of an array to the beginning of the following array. This was achieved by using a user-defined overlap length for both arrays, and within the overlap the array amplitudes were averaged using linear weighing in the range [0,1]. The tapering handled the problem of discontinuity of merely concatenating the signals and non-zero mean resulting from offsetting the signals. The addition of an artifact and a step change in the heart rate can also be applied to the longitudinal signal. However, a change in the signal waveform characteristics is not readily available. This could be added by creating gradually changing waveform characteristics over a certain period of time.

\section{Results}
\label{sec:experiments}

\begin{figure}[]
\centering
\includegraphics[width=0.45\textwidth
]{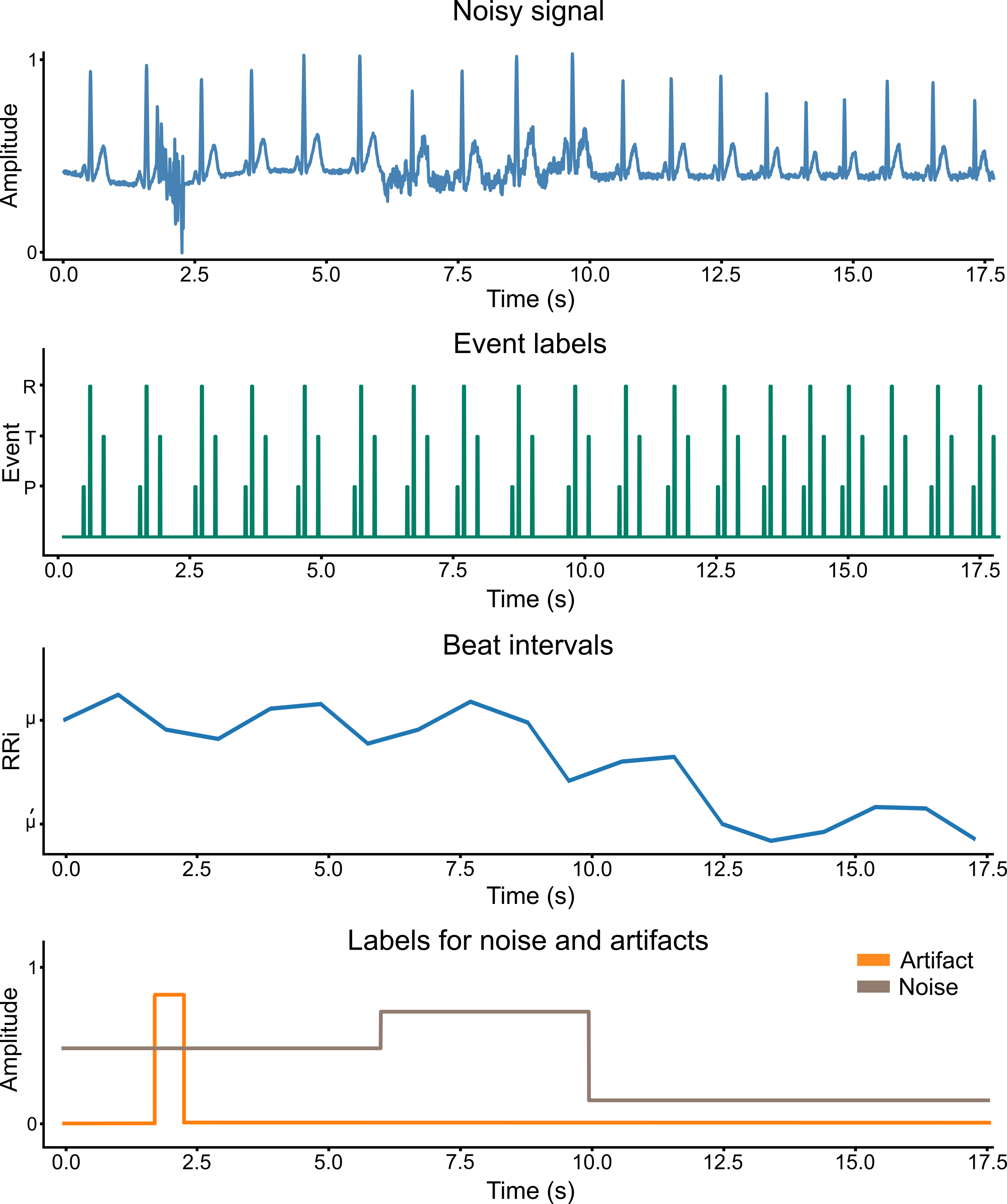}
\caption{Example of a longitudinal signal with three different noises. There is also added artifact and step change in the beat intervals. Labels for noise, artifact and signal events (P, R and T) and beat intervals are also shown in the figure.}
\label{fig:long}
\end{figure}

\begin{figure}[]
\centering
\includegraphics[width=0.45\textwidth
]{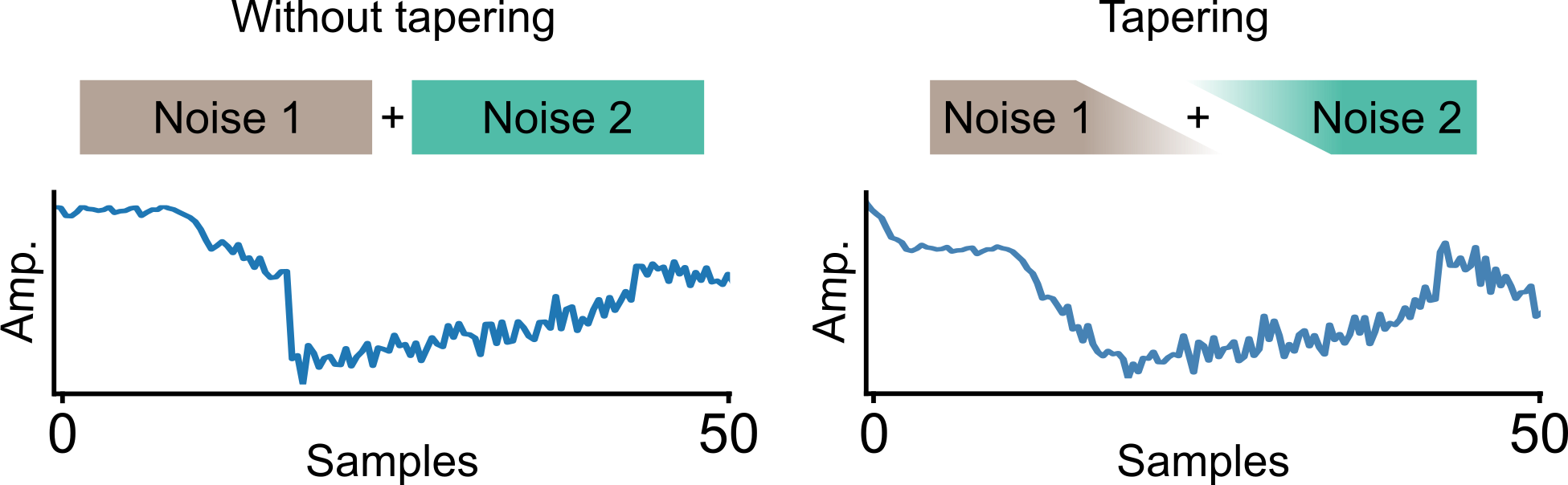}
\caption{Effect of the concatenation of the noise realizations with tapering. }
\label{fig:taper}
\end{figure}

\begin{figure*}[thpb]
    \centering    \includegraphics[width=1.0\textwidth]{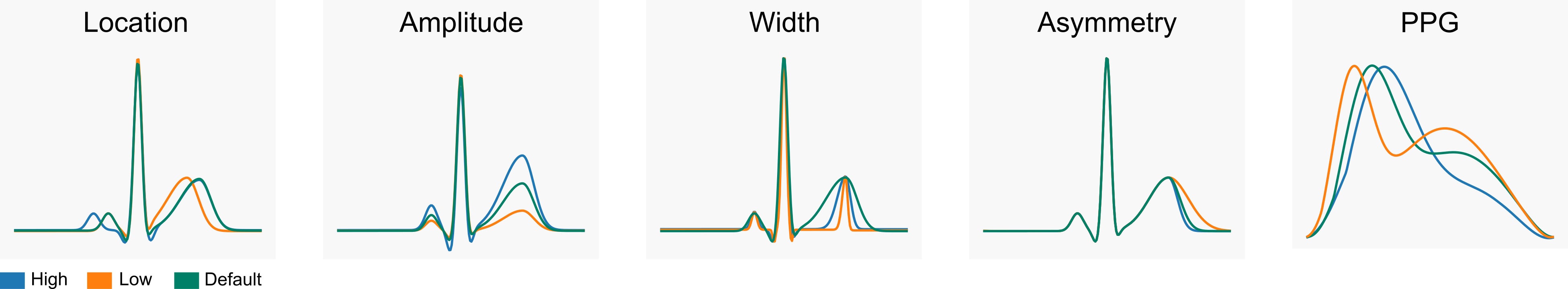}
    \caption{Effect of randomization on ECG and PPG. Location, amplitude, width and asymmetry shows separately the difference in ECG between low and high values on each of the wave parameters. PPG is randomized jointly and the PPG figure shows how the low and high values change the signal.}
    \label{fig:randomization}
\end{figure*}

\subsection{Model capabilities}

The model is capable of producing ECGs and PPGs with various types of noise and artifacts of any duration. An example of a generated long-term ECG that exemplifies the model capabilities is shown in Fig. \ref{fig:long} where; i) different kinds of noises (noise from model PSDs with different parameter values in this case)  were concatenated using the weighted tapering for smooth transition before combining it with the noise-free synthetic ECG signal, ii) a random artifact was added to the signal, iii) heart rate was increased by a step change in the beat intervals, and iv) labels for the noise level, artifact and signal events (P, R and T waves for ECG) were provided. 

A more detailed example of the tapering when concatenating two noise arrays is shown in Fig. \ref{fig:taper}. The discontinuity between the two noise arrays is removed by tapering the ends of the arrays before concatenating them. Similarly, a more detailed example of the step change in the heart rate is shown in Fig. \ref{fig:long}. Using a sigmoid function \eqref{sigmoid}, the step change gradually increases the heart rate by decreasing the mean of RR intervals to the determined mean heart rate.  

\begin{figure}
    \centering
    \includegraphics[width=0.45\textwidth]{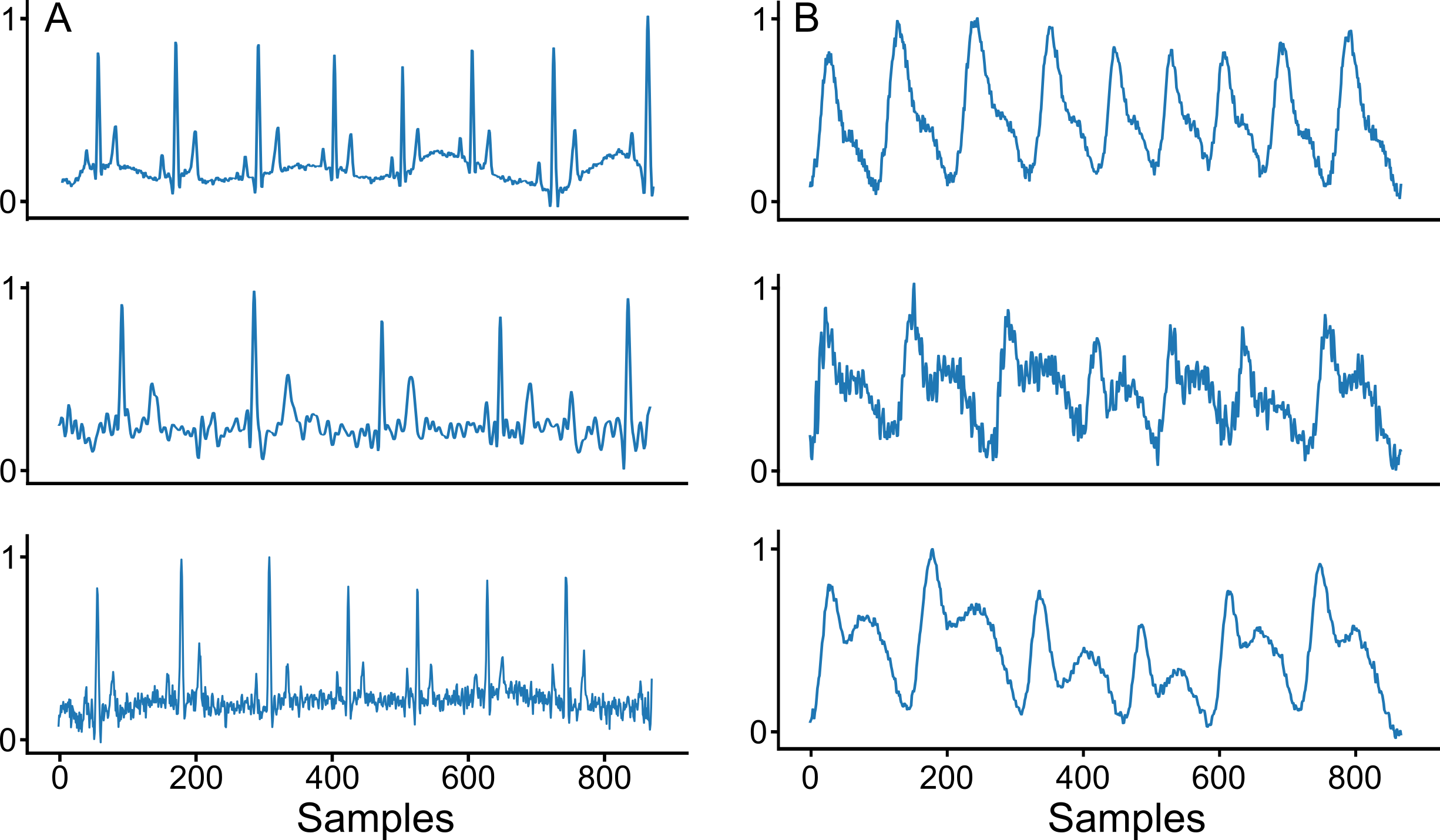}
    \caption{Examples of randomized ECG (A) and PPG signals (B).}
    \label{fig:rand_sig}
\end{figure}

The model is capable of generating a vast amount of signals with randomly varying characteristics for training deep learning models. The effects of randomization limits to the generated biosignals is shown in Fig. \ref{fig:randomization}. Regarding ECG, randomization limits for each wave parameter are shown separately as they are randomized independently. For PPG, the parameters are randomized jointly so only one figure with low, high and default values is shown. Examples of randomized signals are shown in Fig. \ref{fig:rand_sig}. The randomization limits for signal generation can be controlled and changed. This type of a scheme, referred as domain randomization, has been shown to be effective scheme in neural network training \cite{kaisti2023domain}.

\subsection{Visual Validation}

The presented model was validated by comparing the synthetic PPG and ECG signals to measured signals. Fig. \ref{fig:synt_meas} shows an example comparison between one heart cycle (a), noisy ECG (b), and respiratory modulation of the heart rate (c). Synthetic and measured heart cycles align well with similar amplitudes, widths, and locations of each wave. In addition, in the ECG the asymmetry of the T wave is present in both signals. The noisy signal is realistic when compared to ECG signal measured during walking. Respiration modulates the amplitudes of the waves and the RR intervals in a similar way in both signals. The amplitude of the S wave is increased in the measured signal as well as in the synthetic signal. Also the generation of the time realization from the noise PSDs was validated by comparing the original PSD and the PSD computed from converted time realization. Figure of the validation of the generation of the noise and the PSDs of the real noise signals are shown in Fig. \ref{fig:psd_time}. 

The model also produces long-term correlation in the beat intervals that have been shown to be present in healthy awake subjects \cite{kantelhardt2003transiencorr}. Results of the DFA of the long term correlation are shown in Fig. \ref{fig:dfa} where the highest correlation was achieved with parameter \textit{a} value as 1.02 which is used in the model as a default. In addition to the long-term correlation, randomization and step change in heart rate, variation to the beat intervals can also be created by giving the model recorded beat intervals with atrial fibrillation as shown in supplementary figure in Fig. \ref{fig:examples}J.  

More examples of ECG and PPG signals are shown in Fig. \ref{fig:examples} further exemplifying the versatility of the model.

\subsection{Application use cases}
\label{sec:application}

This biosignal model can be used to generate training data that leads to improvements in ECG and PPG deep learning models. We provide some examples below:

\subsubsection{ECG heart rate detection}
Two neural networks, a long short-term memory (LSTM) and dilation convolutional neural network (CONVNET) were trained to detect R peaks as described in \cite{kaisti2023domain}. Training data came from either synthetic ECG signals or real ECG signals from MIT-BIH arrythmia database \cite{moody2001impact, goldberger2000physiobank} augmented with noise from  MIT-BIH Noise Stress Test database \cite{moody1984bih}. We used the ECG-GUDB database \cite{howell2018high} and Computing in Cardiology 2017 single atrial fibrillation database (Cinc2017-AF) \cite{clifford2017af} to test the models. 

The LSTM model consisted of two bidirectional LSTM layers with 64 units and a dense layer with sigmoid activation. The CONVNET model consisted of 7 convolutional layers with dilation rate doubling in each layer from 1 in first layer to 64 in last layer and kernel size of 3.

For the exercise dataset (ECG-GDB), using synthetic ECG to train the models increased the test F1-score from 0.91 to 0.96 and from 0.83 to 0.98 for the LSTM and the CONVNET, respectively, compared when training with real ECG recordings. For the atrial fibrillation dataset, the F1-scores increased when the models were trained with synthetic ECG instead of real ECG recordings (from 0.85 to 0.93 with LSTM and from 0.81 to 0.93 with CONVNET). All the F1-scores are presented in  table \ref{tab:synt_real_comp}.

\subsubsection{PPG peak detection}
To compare efficacy of synthetic signals in PPG peak detection four different models were tested: three models that used real signals (automatic multiscale-based peak detection (AMPD) \cite{scholkmann2012efficient}, multi-scale peak and trough detection (MSPTD) which is an optimized version of the AMPD algorithm \cite{bishop2018multi}, the HeartPy algorithm \cite{van2019heartpy}) and a convolutional neural network (CNN) using synthetic signals. The CNN is presented in \cite{kazemi2022robust}. All the algorithms were tested with real PPG data measured from 20 subjects while sitting, walking or moving hand. More detailed description is presented in \cite{sirkia2024wearable}. The CNN model was trained with synthetic PPG signals and was found to perform the best with overall F1-score of 0.81 compared to the second best with F1-score of 0.74. 

\subsubsection{ECG and PPG segmentation}
A U-Net \cite{ronneberger2015u} trained with synthetic signals and modified to 1-dimensional signals \cite{renna2019deep}, was used to segment each sample of ECG and PPG signals to P and T waves and QRS complex and to systole and diastole, respectively. The U-net was trained using 2000 15 seconds long synthetic ECG or PPG signals. 12 real ECG signals from MIT-BIH Arrhythmia Database P-Wave Annotations dataset \cite{marvsanova2018automatic, goldberger2000physiobank} and six real PPG signals from Pulse Transit Time PPG dataset \cite{mehrgardt2022pulse, goldberger2000physiobank} were used as a test set. One of the PPG signals was measured during walking and others while sitting. The test signals were manually annotated. PPG was segmented with accuracy of 84 \% and area under ROC curve (AUC) of 0.95 and ECG was segmented to four classes (QRS complex, P and T wave and none of these) with accuracy of 77 \% and macro-AUC of 0.82. 

\subsubsection{Quality detector}
A simple three-layered convolutional neural network was trained to classify the quality of ECG signals using only synthetic data, only real data or both real and synthetic data. The quality of the real signals was assessed by visual inspection of the ECG waveform characteristics:
 \begin{itemize}
     \item Level 1 (good signal): P, QRS complex and T waves were recognized by visual inspection.
     \item Level 2 (moderate noise): P or T waves were not recognizable, but the signal can be identified as an ECG and R peaks are recognizable and the heart rate can be calculated from the signal. 
     \item Level 3 (severe noise): The signal can hardly be identified as an ECG and QRS complexes cannot be recognized with certainty.
 \end{itemize}

The quality of the synthetic data was determined by setting limits for the integral of the absolute value of the added randomized noise such that the labels of the generated signals aligned with the quality criteria of the real signals. The limits were adjusted during training based on mislabeled signals in the validation dataset consisting of real ECG signals. Furthermore, if an artifact was added, the quality was set as level 3.

The performance of the models was tested with a dataset consisting of 515 labeled real ECG signals. The quality detector performed the best when trained with both synthetic and real signals with macro-AUC of 0.96 and accuracy of 86 \%. These results are presented in the table \ref{tab:synt_real_comp}.

\begin{table}
    \caption{Application use cases: heart rate detection from ECG and PPG, ECG and PPG segmentation and ECG quality detector.}
    \centering
    \begin{tabular}{cccccc}
    \toprule
    \multicolumn{3}{l}{\textbf{HR ECG} (F1-score)}\\
    \\
    & LSTM & \makecell{Dilation \\CONVNET} & & \\
    \makecell{Real (exercise)} & 0.91 & 0.83\\ \makecell{Synthetic (exercise)} & 0.96 & \textbf{0.98} \\ \makecell{Real (AF)}  & 0.85 & 0.81\\
    \makecell{Synthetic (AF)} & \textbf{0.93} & \textbf{0.93} \\
     \\

    \midrule
    \multicolumn{3}{l}{\textbf{Peak detection PPG} (F1-score)}\\
    \\
    &\makecell{Baseline\\ (AMPD)}  & CNN & MSPTD & HeartPy\\
    Sitting & 0.89  & \textbf{0.96} & 0.89 & 0.93 \\
    Walking & 0.61  & \textbf{0.73} & 0.62 & 0.63 \\
    Hand movement & 0.6  & \textbf{0.71} & 0.62 & 0.63 \\
    All & 0.71  & \textbf{0.81} & 0.72 & 0.74 \\
    \midrule
    \multicolumn{3}{l}{\textbf{Segmentation} ((macro-)AUC/accuracy \%)}\\
    \\    \makecell{ECG\\PPG} & \makecell{ 0.82/77\\ 0.95/84} & \\
    & & &\\
    \midrule
    \multicolumn{3}{l}{\textbf{Quality} (macro-AUC/accuracy \%)}\\
    \\
    Synthetic & Real & \makecell{Real + \\Synthetic}\\
    \\
    0.92/79 & 0.92/70 & \textbf{0.96/86}\\     
    
     & & & &\\
    \bottomrule    
    \end{tabular}

    \label{tab:synt_real_comp}
\end{table}

\section{Discussion}
The presented synthetic biosignal model generates realistic noisy ECG and PPG signals where the characteristics of the signals and the noise can be flexibly varied and randomized to conveniently produce large datasets for various data driven applications. 

As shown in subsection \ref{sec:application}, one of the benefits of this model is automated and accurate labeling. With quantitative noise labels, information regarding signal quality, which is usually lacking from real measurements, the synthetic signals can be used to e.g. train anomaly detectors or quality detectors \cite{syversen2024}. Cardiac cycle segmentation and peak detecting algorithms can also benefit from the model, as it incorporates the labeling of other signal events such as P and T waves. The model could be used to create a standardized test sets for bench-marking various algorithms.

Neural networks require vast amount of training data, and open datasets with proper labels are typically scarce and expensive to collect. By adding synthetic signals to the training set the size of the dataset can be significantly increased. In addition, the synthetic data, by itself or when augmented into existing dataset, can increase variation in the training set, which can improve performance and generalizability of the model. Besides effortless collection, synthetic signals generated with a parametric model do not have any privacy issues. 

Pulse arrival time (PAT) or pulse transit time (PTT) has been intensively studied for cuff-less blood pressure estimation \cite{mukkamala2015toward,panula2022advances}. Since the model can generate several time synchronized signals of either modality, with a given delay between them, the model can be used to simulate signal pairs with a defined PAT and PTT. The model also allows a step type change in the PAT or PTT values during a simulated recording.

The presented biosignal model could be further developed to allow even more options to the signal generation. By adding an option to gradually change the signal parameters, the model could more accurately represent arrhythmia by allowing the parameters to vary along with the irregularity of the beat intervals. 

\section{Conclusion}
We extended the previously presented parametric ECG model to a comprehensive biosignal framework with two signal modalities, ECG and PPG, included flexible noise generation and added domain randomization scheme. In addition to generating the signals itself, the model also produces quantitative noise quality and artifact labels and signal event labels for P and T waves and QRS complex (ECG) and feet and peaks (PPG). The generated synthetic signals could be used for example in development of anomaly and peak detectors, training various deep learning models, bench-marking detection algorithms and educational purposes. 

\appendices
\section{}
\subsection{Summary of model parameters}
Summary of the model parameters is shown in table \ref{tab:modpar}.

\begin{table}[h]
    \centering
    \captionsetup{labelsep=newline, textfont=sc}
    \caption{Model parameters}

    \setlength{\tabcolsep}{5pt}
    \begin{tabular}{c|cccc}
        \toprule
        \textbf{Waveform}    & $d$ & $a$ & $w$ & $m$ \\
        &&&&\\
        \textbf{ECG}&&&&\\
        P & $[-0.18, -0.12]$ & $[0.05, 0.2]$ & $[0.065, 0.085]$ & \\
        Q & $[-0.05,-0.03]$ & $[-0.2, -0.05]$ & $[0.03, 0.08]$ & \\
        R & $0$ & $[0.8, 1.2]$ & $[0.06, 0.085]$ & \\
        S & $[0.03, 0.05]$ & $[-0.2, -0.05]$ & $[0.03, 0.08]$ & \\
        T & $[0.2, 0.25]$ & $[0.1, 0.6]$ & $[0.085, 0.21]$ & $[1, 3]$ \\
        &&&&\\
        \textbf{PPG}&&&&\\
        Sys  & $[-0.32,-0.22]$ & $[0.5,1]$ & $[0.5, 0.9]$ & \\
        Dias & $[0.06, 0.16]$ & $[0.5, 0.9]$ &  $[1.7, 2.1]$ & \\
        \\\midrule
        \textbf{Beat} & $\mu$ & $\beta$ & $f_b$  &  \\
        \textbf{interval} & $[0.4, 1.2]$ & $0.1$ & $0.28$ & $\gamma$ \\
        \\
        $\gamma$ & $a$ & $b$ & $\sigma$\\
         & $1.2$ & $0.075$ & $[0.45, 0.55]$\\
        \\\midrule
        \textbf{Noise}  & $\alpha$ & $c$ & $\sigma^2$ & \\
        & $[0.005, 0.25]$ & $[0, 0.15]$ & $[0, 0.1]$ & \\
        \\\midrule
        \textbf{HR step} & $\mu'$  & $l$  & $\tau$ & \\
        & $[0.3,2]$ & $[0,1]$ & $[1,10]$ &  \\
        \\\bottomrule
    \end{tabular}
    \captionsetup{textfont=normalfont}
    \caption*{$d$ = location of the peak relative to phase \\
    $a$ = amplitude of each wave \\
    $w$ = width of each wave \\
    $m$ = asymmetry parameter \\
    $\mu$ = mean beat interval\\
    $\beta$ = breathing coefficient \\
    $f_b$ = breathing frequency \\
    $\gamma$ = long-term component affected by parameters $a$, $b$ and $\sigma$ \\
    $\alpha$ = exponent of 
    $c$ = scalar of $\frac{1}{f}$ noise \\
    $\sigma^2$ = scalar of white noise \\
    $\mu'$ = mean beat interval after the change in HR \\
    $l$ = location of the change relative to the number of beats \\
    $\tau$ = length of the transition}
    \label{tab:modpar}
\end{table}

\subsection{Modified transient correlations parameter}\label{modifiedgamma}
The long term correlation component $\gamma_i$ of beat interval model is calculated as
\begin{align*}
    &\gamma_i = 0.05\sum^i_{j=1}y_j\Theta(k_j+j-i), \text{ where} \\
    &y_i = x_i \sqrt{1 + \frac{b}{k_i}\left(\sum_{j=0}^{k_i-1} y^2_{i-k_i+j}\right)}, \\
    &k_i \sim \left\lfloor \text{Pareto}(6,a)\right\rfloor, \; x_i \sim N(0, \sigma^2) \text{ and} \\
    &\Theta(m)=\begin{cases}
        0, \; \text{for} \; m \leq 0\\
        1, \; \text{for} \; m > 0
    \end{cases}.
\end{align*}

\subsection{Step change}\label{sigmoidoversignal}
Because the sigmoid function \eqref{sigmoid} is bijective, strictly increasing and symmetric with respect to the point $\left(0,\frac{1}{2}\right)$, for every threshold $\delta \in \left(0,\frac{1}{2}\right) \; \exists x_\delta>0$, such that
\begin{align*}
    S(x) &< \delta, \; \forall x<-x_\delta \text{ and}\\
    S(x) &> 1-\delta, \; \forall x>x_\delta.
\end{align*}
Hence,
\begin{align*}
    & \theta_i=\mu+(\mu'-\mu)S\left(\frac{i-d}{\tau}\right)+\epsilon_i, \quad i \in \mathbb{N}_0\\
        \sim&
        \begin{cases}
            \mu + \epsilon_i, &\forall i<d-x_\delta\\
            \mu+(\mu'-\mu)S\left(\frac{i-d}{\tau}\right)+\epsilon_i, \; &\text{when } d-x_\delta\leq i \leq d+x_\delta\\
            \mu'+\epsilon_i, &\forall i>d+x_\delta
        \end{cases}.    
\end{align*}

\subsection{Phase function in relation to time}\label{xoftime}
The function of time,
\begin{equation*}
    x(t)=\begin{cases}
    \frac{t}{\theta_1},\; &\text{when } t\in[0,\theta_1]\\
    \frac{1}{\theta_k}\left(t-T_{k-1}\right),\; &\text{when } t \in \left( T_{k-1},\: T_k\right], \; \forall k\geq2
    \end{cases},
\end{equation*}
where $T_{k}=\displaystyle\sum_{i=1}^{k}\theta_i$ is the cumulative sum of the beat intervals $\theta_i$, determines the angular velocity of the phase function $\phi$.

\subsection{Complete signal}\label{sumofwaves}
The function $g(\phi; j)$ is the derivative of the Gaussian function for each wave $j$, and these derivatives are summed:
\begin{align*}
    ecg &= \displaystyle\sum_{j \in \{P,\:Q,\: R,\:S,\:T\}} g(\phi; j) \text{ and}\\
    ppg &= \displaystyle\sum_{j \in \{Sys,\: Dias\}} g(\phi; j), \text{ where}\\
    g(\phi; j) &= -\frac{2\pi m_j a_j (\phi+d_j\pi)}{w_j^2}\e{-\frac{m_j(\phi+d_j\pi)^2}{2w_j^2}}.\\
    ECG &= \int ecg\: dt, \text{ and}\\
    PPG &= \int ppg\: dt.
\end{align*}
Ergo, the final pure signal is the derivative, $ecg$ or $ppg$, integrated over time, therefore considering the effect of beat interval length.

\section*{Acknowledgment}
This work was supported by Research Council of Finland project CLISHEAT (352893). 

\section*{Code availability}
The computer code of the model is available in \url{https://github.com/UTU-Health-Research/framework_for_synthetic_biosignals}. 

\printbibliography

\include{supporting}

\end{document}

%% file: supporting.tex
\onecolumn
\section*{\large Supporting information:}
\section*{\large Flexible framework for generating synthetic electrocardiograms and photoplethysmograms}
\vspace{2cm}

\renewcommand{\thefigure}{SI \arabic{figure}}
\setcounter{figure}{0}

\renewcommand{\thetable}{SI \Roman{table}}
\setcounter{table}{0}

\setcounter{page}{1}

\begin{figure*}[thpb]
\centering
\includegraphics[width=\textwidth]{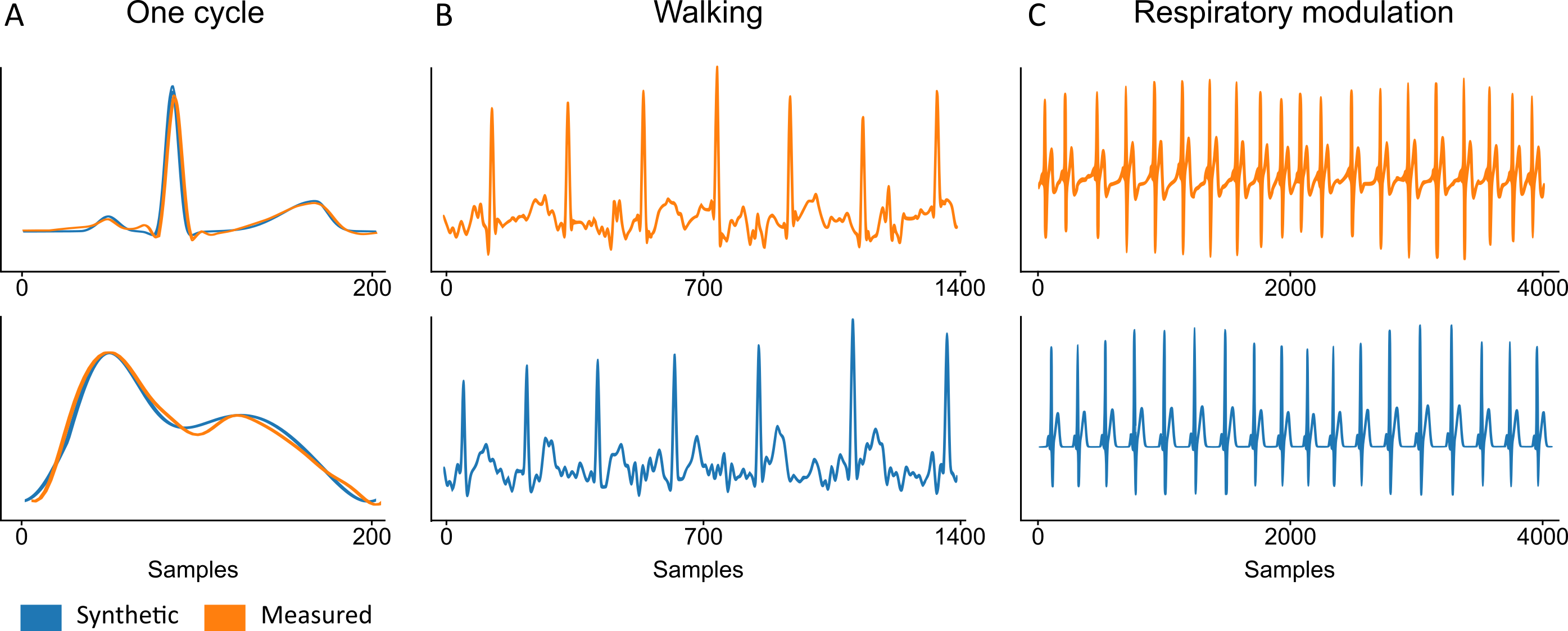}
\caption{Example comparisons of synthetic and measured PPG and ECG signals. Figure a) is comparison of one heart cycle of ECG (upper) and PPG (lower), b) is ECG measured during walking and synthetic ECG with added synthetic noise from walking, and c) is comparison of respiratory modulation of heart rate in ECG.}
\label{fig:synt_meas}
\end{figure*}

\begin{figure*}[thpb]
\centering
\includegraphics[width=\textwidth]{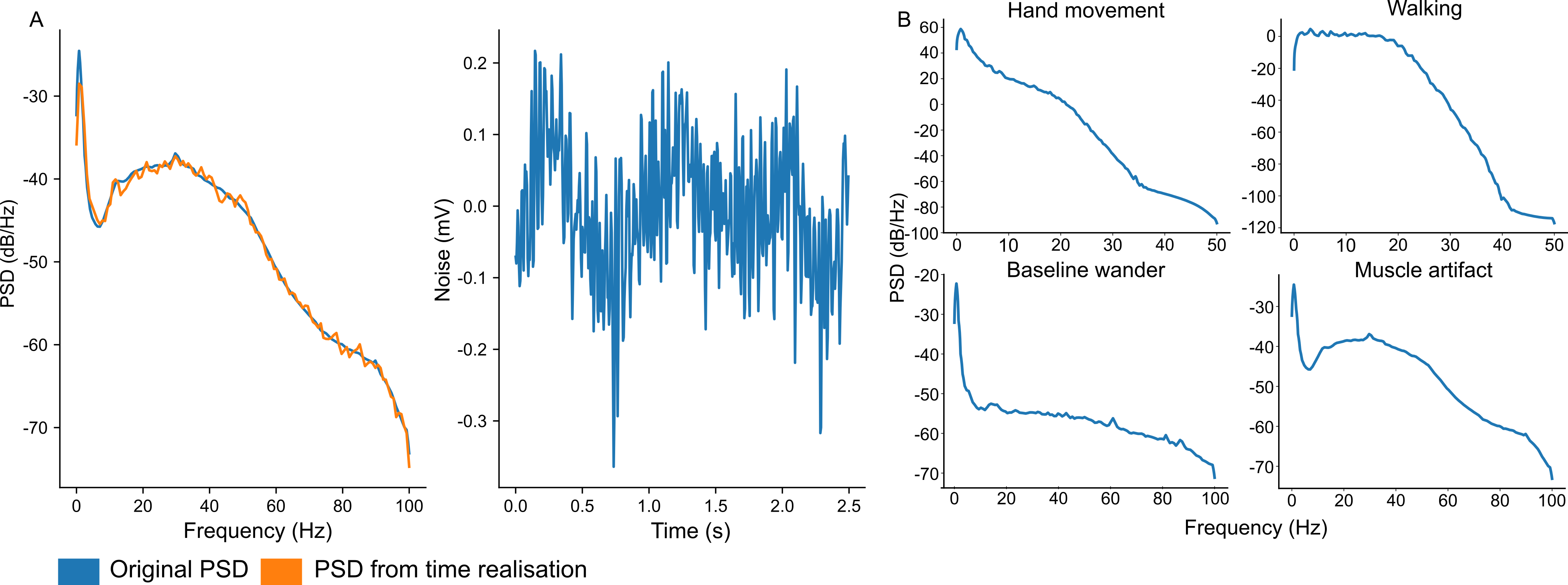}
\caption{Validation of created PSDs and time realisations. Figure A shows the original PSD (left) which was converted to time realisation (right) and then back to PSD (left). Figure B shows PSDs computed with the Welch's method from noise measurements (hand movement and walking from PPG signals and baseline wander and muscle artifact from ECG signals).}
\label{fig:psd_time}
\end{figure*}

\label{dfa}
\begin{figure}[thpb]
\centering
\includegraphics[width=0.9\textwidth]{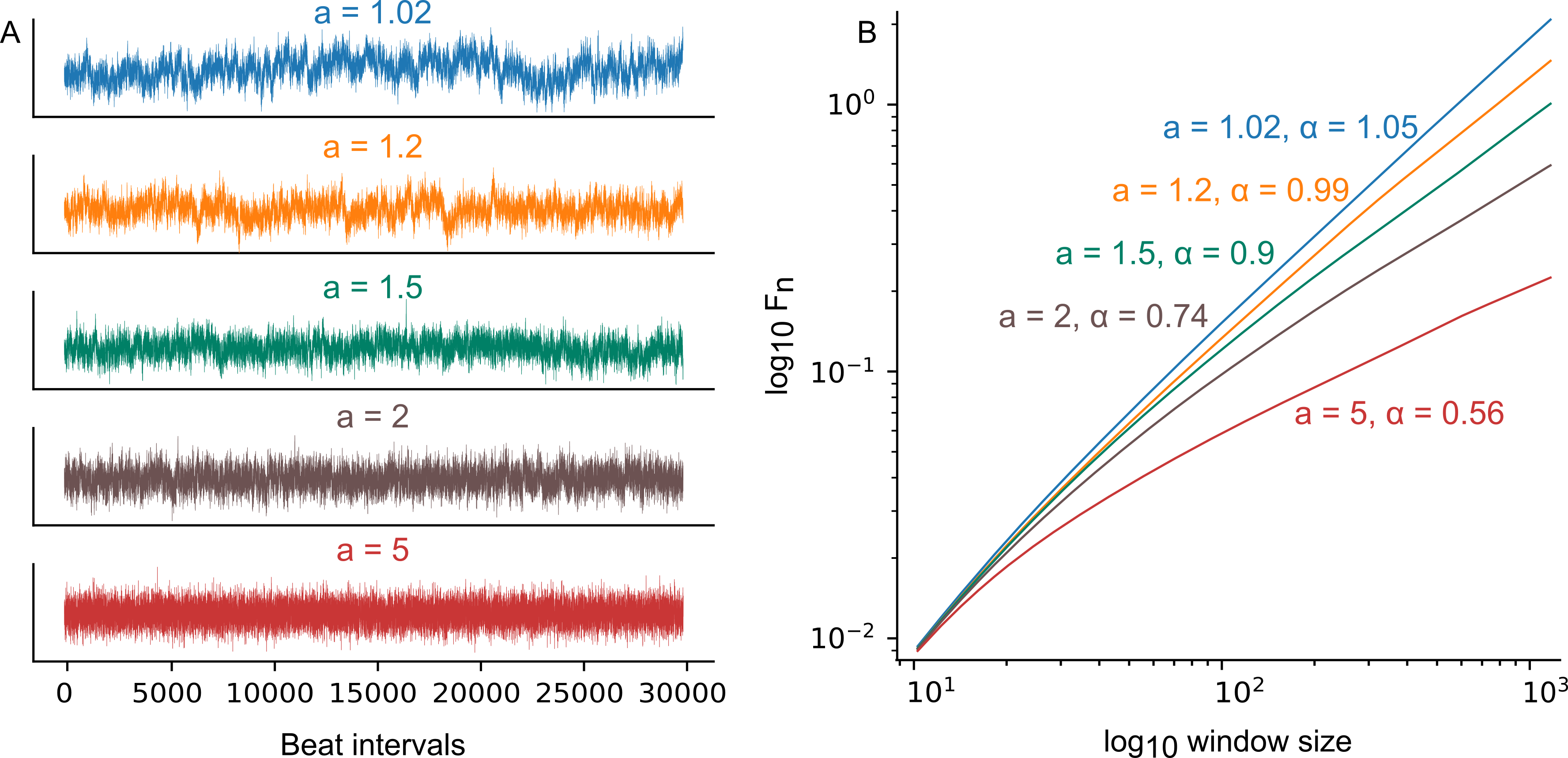}
\caption{Results of detrend fluctuation analysis (DFA) of series of 100,000 generated beat intervals, where the value $\alpha \approx 1$ corresponds to $1/f$ noise and long term correlations observed in awake healthy subjects, and $\alpha \approx 0.5$ corresponds to white noise and uncorrelated beat intervals observed in healthy subjects during deep sleep \cite{kantelhardt2003transiencorr}. Figure A shows 30 000 of the beat intervals used for the DFA and figure B the results of the DFA ($F_n$ is total fluctuation value).}
\label{fig:dfa}
\end{figure}

\begin{figure*}[thpb]
\centering
\includegraphics[width=\textwidth]{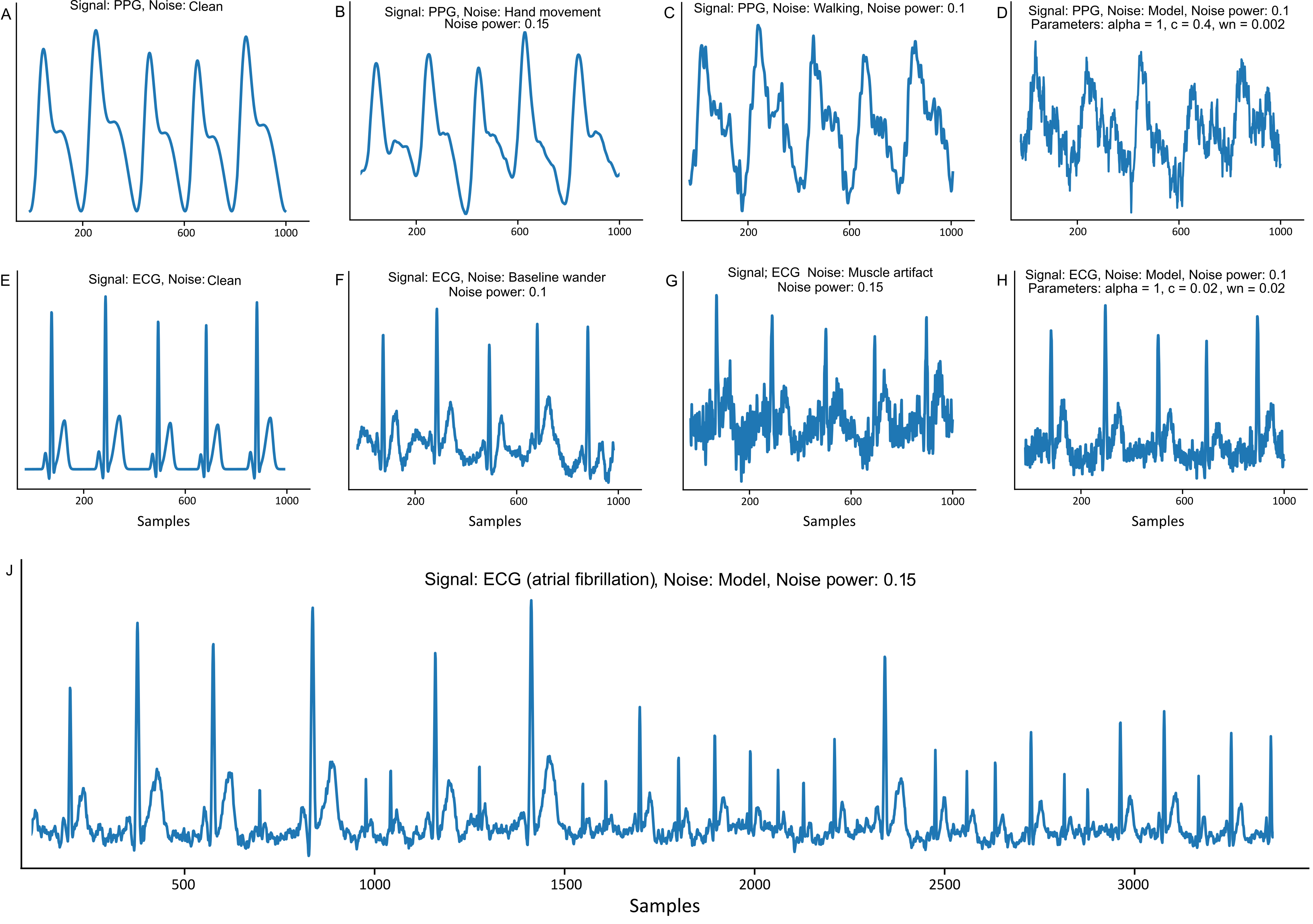}
\caption{Examples of clean and noisy ECG and PPG signals and ECG with atrial fibrillation.}
\label{fig:examples}
\end{figure*}